\documentclass[journal]{IEEEtran}

\pdfoutput=1

\ifCLASSINFOpdf
  \usepackage[pdftex]{graphicx}
  \graphicspath{{figures/}}
\fi

\usepackage{amsmath,amssymb}

\usepackage{url}

\begin{document}

\onecolumn
\begin{center}

\huge
\textbf{\textcopyright 2018 IEEE}
\linebreak\linebreak
\large
Pre-print of:
\linebreak
C. Henry, S. M. Azimi, N. Merkle, "Road Segmentation in SAR Satellite Images with Deep Fully-Convolutional Neural Networks", IEEE Geoscience and Remote Sensing Letters, 2018, accepted for publication.
\linebreak
DOI: 10.1109/LGRS.2018.2864342
\linebreak\linebreak
This material is posted here with permission of IEEE.
\linebreak\linebreak
Personal use of this material is permitted.
\linebreak\linebreak
Permission from IEEE must be obtained for all other uses, in any current or future media, including reprinting/republishing this material for advertising or promotional purposes, creating new collective works, for resale or redistribution to servers or lists, or reuse of any copyrighted component of this work in other works, by writing to:
\linebreak
\url{pubs-permissions@ieee.org}
\linebreak\linebreak
By choosing to view this document, you agree to all.

\end{center}
\twocolumn
\newpage

\title{Road Segmentation in SAR Satellite Images with Deep Fully-Convolutional Neural Networks}

\author{Corentin~Henry,~
        Seyed~Majid~Azimi~
        and~Nina~Merkle
        \vspace{-0.8cm}
\thanks{The authors are with the Remote Sensing Technology Institute of the German Aerospace Center (DLR),  Wessling 82234, Germany (e-mail: corentin.henry@dlr.de; seyedmajid.azimi@dlr.de; nina.merkle@dlr.de)}}

\markboth{PRE-PRINT ACCEPTED FOR PUBLICATION IN IEEE GEOSCIENCE AND REMOTE SENSING LETTERS}%
{Henry \MakeLowercase{\textit{et al.}}: Road Segmentation in SAR Satellite Images with Deep Fully-Convolutional Neural Networks}

\maketitle

\begin{abstract}
Remote sensing is extensively used in cartography. As transportation networks grow and change, extracting roads automatically from satellite images is crucial to keep maps up-to-date. Synthetic Aperture Radar satellites can provide high resolution topographical maps. However roads are difficult to identify in these data as they look visually similar to targets such as rivers and railways.
Most road extraction methods on Synthetic Aperture Radar images still rely on a prior segmentation performed by classical computer vision algorithms. Few works study the potential of deep learning techniques, despite their successful applications to optical imagery.
This letter presents an evaluation of Fully-Convolutional Neural Networks for road segmentation in SAR images. We study the relative performance of early and state-of-the-art networks after carefully enhancing their sensitivity towards thin objects by adding spatial tolerance rules. Our models shows promising results, successfully extracting most of the roads in our test dataset. This shows that, although Fully-Convolutional Neural Networks natively lack efficiency for road segmentation, they are capable of good results if properly tuned. As the segmentation quality does not scale well with the increasing depth of the networks, the design of specialized architectures for roads extraction should yield better performances.
\end{abstract}


\IEEEpeerreviewmaketitle

\vspace{-0.5cm}
\section{Introduction}

\IEEEPARstart{T}{he} overall urban growth in the past two decades has led to a considerable development of transportation networks. Such constantly evolving infrastructure necessitates frequent updates of existing road maps.
A wide range of applications are depending on this information, such as city development monitoring, automated data update for geolocalization systems or support to disaster relief missions.
A satellite equipped with a Synthetic Aperture Radar (SAR) can get information on an area's topography. The resulting information is more robust to changes in illumination conditions and color fluctuation with respect to optical imagery. Moreover, SAR sensors can operate independently from weather conditions, and are therefore the sensor of choice to survey regions affected by weather-related disasters.

The extraction of roads in SAR satellite images has been researched for several decades \cite{Tupin1998DetectionExtraction} and is generally addressed in the following manner: road candidates are extracted from SAR images using a feature detector. This initial segmentation is then transformed into a topological graph, where each segment represents a road section. The graph is finally optimized to form a coherent road network, often by applying a Markov Random Field (MRF) \cite{Tupin1998DetectionExtraction, Perciano2011AData} using contextual information from the SAR image to reconnect loose segments and correct the overall network structure. Recently, Xu et al. \cite{Xu2017BayesianImages} proposed a Conditional Random Field (CRF) model capable of jointly extracting road candidates and applying topological constraints. This end-to-end scheme reduced the inevitable performance loss occurring when separately extracting road priors and constructing a road network graph. These methods all rely on an efficient road candidate extraction algorithm and most of them entrust this task to traditional computer vision algorithms. To date, few works study the potential of the recent advances in deep learning in the context of road segmentation.

Deep Convolutional Neural Networks (DCNNs) first demonstrated unmatched effectiveness in 2012 on the ImageNet classification challenge and their performance has been improving at a fast pace ever since, receiving a lot of attention from the computer vision community. However, unlike the medium-sized images used in classification competitions, the aerial images used in remote sensing often cover hundreds of square kilometers. Today, Fully-Convolutional Neural Networks (FCNNs) are the most successful method to perform pixel-wise segmentation on large-scale images. Given an input image, they produce an identically-sized prediction map. Introduced in 2015 with FCN8s \cite{Long2015FullySegmentation}, FCNNs allowed the establishment of new states-of-the-art in semantic segmentation of aerial optical images \cite{Zhang2018RoadUnet} and were successfully applied to satellite SAR images \cite{Yao2017SemanticPairs}.

\begin{figure}[t!]
\centering
\includegraphics[width=0.5\textwidth]{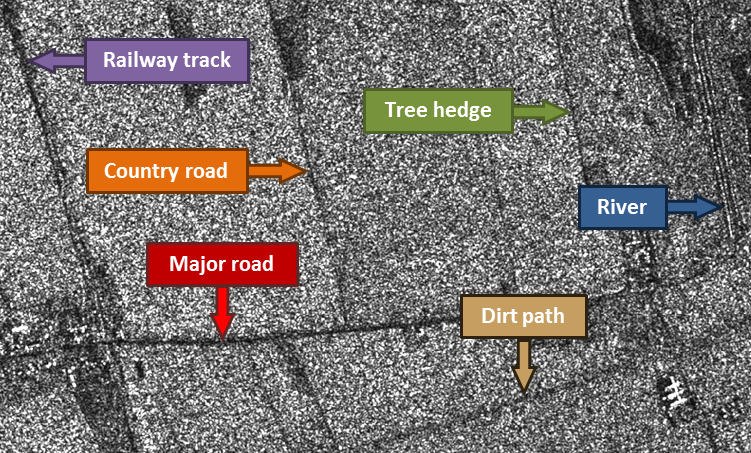}
\vspace{-0.6cm}
\caption{SAR image sample showing that objects of different natures can look very similar. A segmentation model must learn to distinguish all kinds of roads from railway tracks, tree hedges and rivers.}
\label{fig:objects_likeness}
\vspace{-0.6cm}
\end{figure}

In \cite{Yao2017SemanticPairs}, Yao et al. use off-the-shelf pre-trained FCNNs on SAR images to classify buildings, landuse, bodies of water and other natural areas. They report good segmentation results for the landuse and natural classes but unsatisfactory results for buildings, showing a striking performance contrast between larger and smaller objects. As roads are thin objects by nature, it becomes evident that FCNNs models must be specifically adjusted for our task. Starting from another perspective, Geng et al. successively proposed two methods for land cover classification, including roads. In \cite{Geng2017DeepClassification}, they emphasize low-level features in SAR images using traditional computer vision techniques on top of which they train a stack of auto-encoders. In \cite{Geng2018SarNetworks}, they further improve their results by using Long-Short-Term Memory units (LSTM) to transform the 2-D information contained in the image into 1-D information fed to auto-encoders. They report around 95\% overall accuracy across several areas totaling 21 km\textsuperscript{2}, however the road class accuracy remains invariably behind the accuracy of all other classes by 10\% to 20\%.

Roads are difficult to identify even in high resolution SAR images. They can often be confused with other targets such as railway tracks, rivers or even tree hedges, as illustrated in Fig. \ref{fig:objects_likeness}. Identifying roads often involves the opinion of an expert, but deep learning proved it could deal with such delicate study cases, motivating the thorough assessment of the potential of some powerful FCNNs on the task at hand. The success of this initial experiment would open new prospects in the future, like semi-automated annotation of SAR images, which could prove much faster and more reliable than fully-manual annotation. Time-critical missions such as disaster relief would particularly benefit from a steep increase in the speed of satellite data analysis.

This letter presents the evaluation of three FCNNs for road segmentation in high resolution SAR satellite images: FCN-8s \cite{Long2015FullySegmentation}, Deep Residual U-Net \cite{Zhang2018RoadUnet} and DeepLabv3+ \cite{Chen2018EncoderdecoderSegmentation}. Crucial adjustments are made in the training procedure to improve the base performance of the FCNNs, with a class-weighted Mean Squared Error (MSE) loss and a control parameter over the spatial tolerance of the models.
The evaluation is performed on several custom datasets, whose design is critical to the success of the method and is therefore detailed. Unlike Yao et al. in \cite{Yao2017SemanticPairs}, we set aside the OpenStreetMap (OSM) data due to the lower geo-localization accuracy compared to SAR data. Unlike previous works, we manually label every single road from the most visible highways to the less distinguishable dirt paths.
We obtain good qualitative results and satisfying quantitative results, thus demonstrating the effectiveness of well-fitted FCNNs as road candidate extractors in SAR images.

\section{Method}
\subsection{Segmentation with Fully-Convolutional Neural Networks} \label{section:segmentation_fcnn}

FCNNs are currently the most successful methods for pixel-wise segmentation, and are especially convenient for large scale image processing.
As they can deal with images of any size, they can take into account a wider context when trying to identify objects.
They owe this flexibility property to their adaptive bottleneck layers, connecting the two key components of the network.
The first element, a DCNN encoder, analyzes the images and outputs a cluster of predictions. The image data is gradually down-sampled, proportionately becoming more meaningful.
The second element, a decoder, applies up-sampling operations to restore the spatial properties of the predictions until the predictions share the same size as the input image. It is often done using bilinear interpolation or fractionally strided convolutions, also called deconvolutions \cite{Zeiler2010DeconvolutionalNetworks}.
For classification tasks, the DCNN output is classified by fixed-size fully-connected layers, the network's bottleneck, imposing a maximum input size upstream.
For segmentation tasks, FCNNs remove this input size constraint by replacing the fully-connected layers by convolutional layers.

We implement three substantially different FCNNs.
The first one is FCN-8s \cite{Long2015FullySegmentation} with a VGG-19 backbone \cite{Simonyan2015VeryRecognition}, the first of all FCNNs which was successfully applied to a wide variety of computer vision tasks. In FCN-8s, two skip connections fuse the high resolution information from early VGG19 layers into the up-sampling process, thus improving the spatial accuracy of the resulting segmentation. To increase its training speed, we add a batch normalization step between each convolutional layer and ReLU activation, as well as after each deconvolutional layer. We use it to set the baseline performance for comparison with more recent architectures.
The second one is Deep Residual U-Net \cite{Zhang2018RoadUnet} which demonstrated a great segmentation performance on the Massachusetts roads dataset \cite{Mnih2013MachineLabeling}. Its overall architecture is similar to FCN-8s, although entirely symmetrical with a skip connection fusing each block of the encoder into the corresponding block of the decoder. Its backbone uses residual units \cite{He2016DeepRecognition} which let the input image data flow through the whole network. Propagating this information helps the network learn complex patterns more efficiently, and its application on SAR imagery could help reduce the impact of the speckle.
The third one is DeepLabv3+ \cite{Chen2018EncoderdecoderSegmentation}, one of the most recent architecture for semantic segmentation. Its Xception backbone also uses residual connections, but the network is much deeper with 65 non-residual layers compared to Deep Residual U-Net's 15. Using dilated convolutions, DeepLab can leverage a larger context and better recognize targets from cluster, which should prove valuable for applications to SAR imagery.

\subsection{Adjusting the FCNNs for road segmentation} \label{section:adjustingFCNN}

Roads appear as thin objects in SAR images and are likely outweighted by clutter, especially outside cities. We take some necessary steps to limit the class imbalance during training.
A similar problem in the case of sports field lines extraction is addressed in \cite{Homayounfar2017SportsModels} by tracing thick labels in the ground truth. In our case, it means that the labels must exactly cover the road outlines and embankments, insofar as they are visible in the SAR images. Pixels labeled as roads are set to a value of 1 and background pixels to 0 in the ground truth. In addition, we introduce a spatial tolerance parameter $t_{max}$ operating as follows. The value of background pixels located at a distance $t \leq t_{max}$ to the nearest pixel labeled as road is redefined as: $1 - \frac{t}{t_{max} + 1}$. The resulting ground truth is a smooth target distribution centered around the road labels, similar to what Luo et al. proposed in \cite{Luo2016EfficientMatching}. Varying $t_{max}$ allows controlling the tolerance of the training towards spatially small mistakes. Note that when referring to a binary ground truth (2 classes) in the following parts, we assume $t_{max}=0$.

As a consequence, the task changes from a binary classification to a binary regression: instead of predicting each pixel as either road or background, the network weighs how much each pixel is likely to be a road. We make the following changes to adapt the FCNNs: the final activation on the logits is changed from a softmax to a sigmoid function, and the cross-entropy loss is replaced by a Mean Squared Error loss (MSE).

Eigen and Fergus \cite{Eigen2015PredictingArchitecture} also tackle the class imbalance issue by reweighting each class upon the loss calculation. The loss for each pixel prediction is multiplied by a coefficient inversely proportional to the frequency of its true class in the ground truth. However, the median class frequency is used to compute these coefficients, which is irrelevant in our case since we only have two classes. Therefore, we set the background class weighting coefficient to 1 and test several road class weighting coefficients taken in the interval $W=[1,1/f_{road}]$ where $f_{road}$ is the ratio of road pixels over total pixels in the entire ground truth.
The MSE loss thus becomes:

\vspace{-0.2cm}
\begin{equation}
Loss_{MSE}(Y_{tol},\hat{Y})=\frac{1}{N}\sum_{i=1}^{N}{w_i(y_i - \hat{y}_i)^2}
\end{equation}
where $y_i$ is the value in the tolerant ground truth $Y_{tol}$ and $\hat{y}_i$ is the sigmoid value in the predictions $\hat{Y}$, for pixel $i$. The number of pixels in the image is given as $N$ and the loss weighting coefficient $w_i$ for pixel $i$ is defined as:
$$
w_i=
\begin{cases}
	\lambda	& \quad \text{if pixel } i \text{ is 1 (road) in } Y_{bin}\\
    1		& \quad \text{if pixel } i \text{ is 0 (background) in } Y_{bin}
\end{cases}
$$
where $\lambda$ is a fixed value taken from the interval $W$ and $Y_{bin}$ is the binary ground truth.

\subsection{Applying pre- and post-processing} \label{section:prepostprocessing}

We study the effect of two operations commonly applied to similar tasks: Non-Local (NL) filtering of SAR images \cite{Buades2005ADenoising} and segmentation post-processing with Fully-connected Conditional Random Fields (FCRFs) \cite{Krahenbuhl2011EfficientPotentials}.
NL filtering improves the overall feature homogeneity in SAR images, often mitigating the negative effects of speckle noise.
FCRFs have been very successful in improving the consistency of FCNN segmentation maps, especially by refining the borders between object regions. They optimize an energy function combining two spatial- and color-based correlation potentials in order to remove inconsistent predictions and refine correct ones. They can be extremely valuable since road segmentation is very sensitive to object smoothness.

\section{Experiments} \label{section:experiments}

\subsection{Experimental procedure} \label{section:experimental_procedure}

\textbf{Dataset:} To the best of our knowledge, there is no publicly available dataset suitable for our study case. We created our own dataset using high resolution TerraSAR-X images acquired in spotlight mode (see table \ref{table:dataset}). We identified the roads as either major roads, country roads or dirt paths with the help of Google Earth optical images. Each road type was assigned a specific label thickness, best matching their respective outline thickness overall. The masks for all road types were merged into a binary ground truth, which was then smoothed as explained in section \ref{section:adjustingFCNN}.
However, manually labeling roads in urban areas was impractical: most objects were either difficult to distinguish or very similar to roads but of a different nature such as building edges. For this reason, we selected regions with fairly dense road networks and very few cities, from which we removed all urban areas. We used a land segmentation map\footnote{\url{http://land.copernicus.eu/pan-european/corine-land-cover/clc-2012}} to delimit and mask out most cities, then manually removed the remaining ones.

\begin{table}[t!]
\centering
\caption{Metadata of the TerraSAR-X images used in our dataset}
\vspace{-0.2cm}
\label{table:dataset}
\begin{tabular}{| l | c |} 
\hline
\textbf{Lincoln, England} & \\ 
\hline
Size, Ground Sample Distance	& 20480*12288 px, 1.25 m/px \\
Projection Coordinate System	& WGS 84 / UTM zone 30N	\\
Coordinates Top-Left			& [683056.875, 5931158.125]	\\
Coordinates Bottom-Right		& [698416.875, 5905558.125] \\
Reference Time UTC				& 2009-12-27T06:25:21.938000Z \\
\hline
\textbf{Kalisz, Poland} & \\ 
\hline
Size, Ground Sample Distance	& 4000*4000 px, 1.25 m/px \\
Projection Coordinate System	& WGS 84 / UTM zone 34N	\\
Coordinates Top-Left			& [316607.000, 5720181.000]	\\
Coordinates Bottom-Right		& [321607.000, 5715181.000] \\
Reference Time UTC				& 2009-04-12T04:59:32.920000Z \\
\hline
\textbf{Bonn, Germany} & \\ 
\hline
Size, Ground Sample Distance	& 3600*4080 px, 1.25 m/px \\
Projection Coordinate System	& WGS 84 / UTM zone 32N	\\
Coordinates Top-Left			& [356400.000, 5630000.000]	\\
Coordinates Bottom-Right		& [361500.000, 5625500.000] \\
Reference Time UTC				& 2009-01-22T05:51:25.023344Z \\
\hline
\end{tabular}
\vspace{-0.4cm}
\end{table}

\textbf{Training:} We implemented the networks using Tensorflow 1.4 and trained them on a single NVIDIA Titan X Pascal. All networks were trained from scratch, as we noticed a considerable performance drop when using weights pre-trained on ImageNet, certainly due to the different nature of SAR images compared to optical ones. The convolutional weights were initialized with He uniform distributions\footnote{\url{www.tensorflow.org/api_docs/python/tf/keras/initializers/he_uniform}}, the deconvolutional weights with bilinear filters and the biases with zeros. We used an ADAM optimizer with a learning rate of 5e-4 and an exponential learning rate decay of 0.90 applied after each epoch. When not using weights pre-trained on 3-channel RGB images, 1-channel SAR images could be used as input since the weights were initialized accordingly. The area of Lincoln was split into a training and test set as follows: the upper 80\% of the image (16384*12288 px) was used for training, the lower 20\% (4096*12288) for testing. The images from Kalisz and Bonn were used as additional test sets. The input data was normalized and data augmentation was performed on the training set, with patch rotations (0$^{\circ}$, 90$^{\circ}$, 180$^{\circ}$ and 270$^{\circ}$), horizontal and vertical flips. The augmented training set is composed of 12288 patches, referred to as the epoch data.

\textbf{Evaluation metrics:} For the evaluation, the predictions were thresholded at 0.5 to obtain a binary mask, which was then compared to the binary ground truth. We evaluated the performance of our models by computing the Intersection over Union (IoU) ($\frac{TP}{TP+FP+FN}$), the precision ($\frac{TP}{TP+FP}$) and the recall ($\frac{TP}{TP+FN}$), where $TP$, $FP$, $TN$ and $FN$ denote the total number of true positives, false positives, true negatives and false negatives for the road predictions, respectively.
The IoU is a robust metric for segmentation quality assessment since it yields the overlapping ratio between predictions and labels (intersection) over their total surface (union). If the predictions match the labels well and do not extend outside of them, the IoU score will be high.
Coupled with the precision (prediction correctness) and recall (prediction completeness), we can assess accurately the performance of a model.
Although very common in computer vision, the accuracy metric (\(\frac{TP+TN}{TP+FP+FN+TN}\)) is unsuitable for our study case. Since roads make up for around 5\% of the pixels in our ground truth, 95\% of accuracy could mean that only background was predicted.

\subsection{Discussion} \label{section:Discussion}

To setup a baseline performance, we optimize our hyperparameters $t_{max}$ and the loss weighting coefficient on FCN-8s and present the results on the test area from Lincoln (see table \ref{table:results_fcn8s}). We then train a Deep Residual U-Net model and a DeepLabv3+ model using the best parameters found for $t_{max}$ and the weighted loss function, and compare their results with the corresponding FCN-8s model across all our test images.

\textbf{Adapting the spatial tolerance $t_{max}$:} We test the following tolerance values: 0, 1, 2, 4 and 8 px. As anticipated, the greater the tolerance, the better the ground truth coverage (+13\% recall between 0px and 8px of tolerance), at the cost of a larger loss in precision (--17\%).
The best model reaches 44.98\% IoU for $t_{max} = 4 px$. There is a compromise between precision and recall, with a loss below 10\% in precision for a gain of 8\% in recall compared to the model with $t_{max} = 0 px$. We maintain $t_{max} = 4 px$ for the rest of the experiments.

\textbf{Adjusting the loss weighting:} Around 5\% of the pixels in the ground truth are roads (inverse frequency: $1/0.05=20$), therefore we experiment on the following loss weighting coefficients: 1, 2, 4 and 8. Loss weighting induces a maximum gain of 0.48\% IoU with a coefficient of 2, reaching a value of 45.46\%.

\textbf{Applying NL-filtering:} The results become worse when using NL-filtered SAR images, with a considerable decrease of 8\% in IoU. A plausible explanation is that FCNNs natively apply spatial filtering through down-sampling and convolutional filtering, so NL-filtering discards meaningful information.

\textbf{Applying FCRFs post-processing:} Contrary to our expectations, FCRFs fail to improve the connectivity of severed predicted road sections. In our case, the overall result is close to that of an erosion operation, removing not only spurious predictions but also valid ones. Moreover the segmentation is already smooth and regular, limiting the benefits of FCRFs.

\begin{figure}[t!]
\centering
\includegraphics[width=0.5\textwidth]{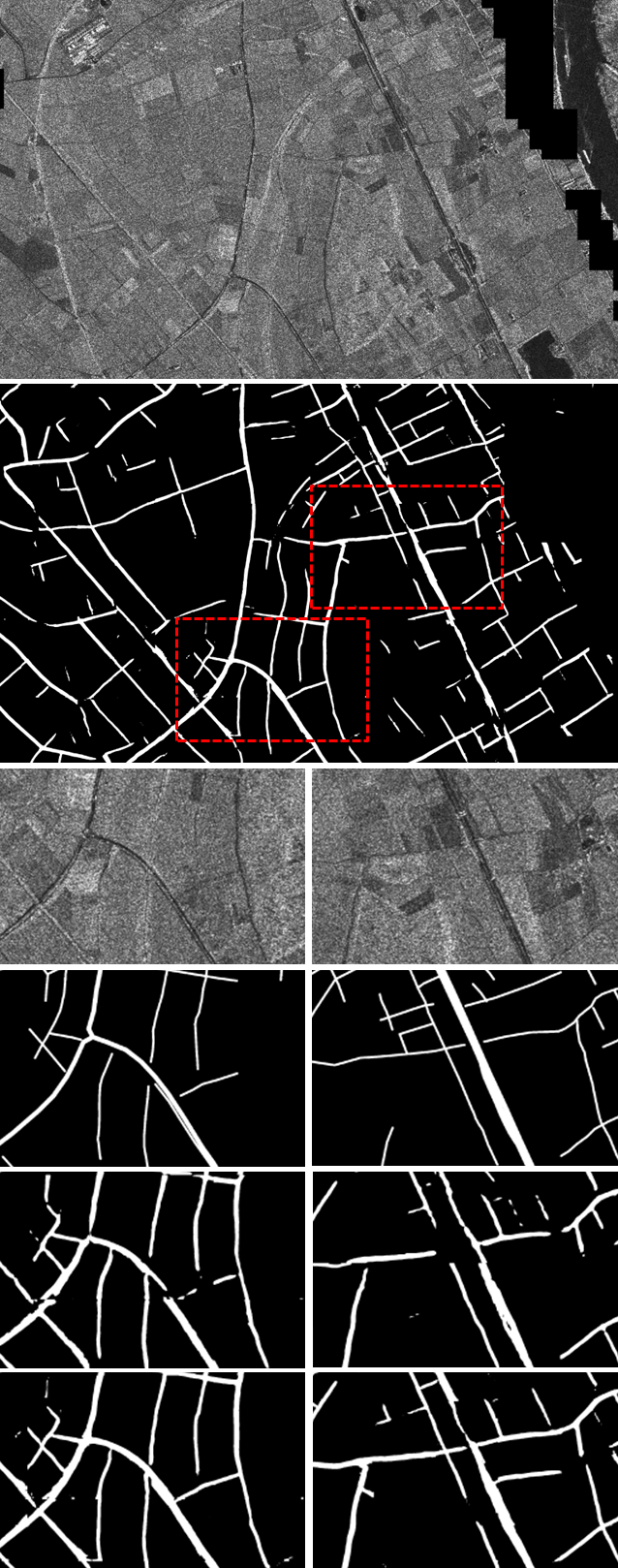}
\vspace{-0.6cm}
\caption{Segmentation results of FCN-8s and DeepLabv3+ over the area of Bonn. Top to bottom: SAR image with masked urban areas, DeepLabv3+ predictions, zoomed samples. In the samples, top to bottom: SAR image, ground truth, FCN-8s predictions, DeepLabv3+ predictions.}
\label{fig:figure_results_bonn}
\vspace{-0.4cm}
\end{figure}

\textbf{Additional FCNNs and test images:} We train a Deep Residual U-Net model and a DeepLabv3+ model with $t_{max} = 4 px$ and a loss weighting coefficient of 2. We report the results over the full test set in table \ref{table:results_iou} to compare their performance to FCN-8s and assess the generalization capacity of each architecture. On the one hand, Deep Residual U-Net shows surprisingly low performance compared to the other models. Like DeepLab, this network propagates the noisy SAR data down to deep layers but, unlike DeepLab, does not have a sufficient depth to abstract from it. On the other hand, DeepLabv3+ and FCN-8s achieve very close scores for all test images. Their performance is moderately reduced (max. --4.73\% IoU) when applied to images from another region with respect to the training area. However, we find out that DeepLabv3+ converges 2.4 times faster than FCN-8s and produces far smoother and less noisy road predictions. A visualization of the segmentation of DeepLabv3+ over the area of Bonn is shown in Fig. \ref{fig:figure_results_bonn}.

\begin{table}[t]
\centering
\caption{Performance of FCN-8s over the test area in Lincoln}
\vspace{-0.2cm}
\label{table:results_fcn8s}
\begin{tabular}{|c c c c c|}
\hline
$t_{max}$	& Loss weight	& IoU		& Precision	& Recall \\ 
\hline
0 px		& 1				& 43.79\%	& \textbf{71.69}\%	& 52.94\% \\
1 px		& 1				& 44.44\%	& 70.68\%	& 54.48\% \\
2 px		& 1				& 44.93\%	& 69.45\%	& 56.00\% \\
4 px		& 1				& 44.98\%	& 62.96\%	& 61.16\% \\
8 px		& 1				& 42.92\%	& 54.72\%	& 66.56\% \\
4 px		& 2				& \textbf{45.46}\%	& 65.34\%	& 59.91\% \\
4 px		& 4				& 45.21\%	& 57.96\%	& 67.27\% \\
4 px		& 8				& 43.73\%	& 51.13\%	& \textbf{75.17}\% \\
\hline
\end{tabular}
\vspace{-0.4cm}
\end{table}

\begin{table}[t]
\centering
\caption{IoU scores of the best models over three test areas}
\vspace{-0.2cm}
\label{table:results_iou}
\begin{tabular}{|c c c c|}
\hline
Area 	& FCN-8s 	& Deep Res. U-Net 	& DeepLabv3+ \\ 
\hline
Lincoln	& 45.46\%	& 40.18\%			& \textbf{45.64}\% \\ 
Kalisz	& 43.85\%	& 27.31\%			& \textbf{44.66}\% \\
Bonn	& \textbf{42.57}\%	& 35.90\%			& 40.91\% \\
\hline
\end{tabular}
\vspace{-0.4cm}
\end{table}

\textbf{Limits of the method:} Our models achieving the best results had difficulties generalizing over a wide variety of patterns, predicting unexpected objects like mounds and forest borders and missing many roads, mostly the less visible ones. A visual inspection of the results shows the limits of the proposed annotation scheme, as the label thickness based on road types does not reflect the actual thickness of many roads. Many prediction failures are due to this shortcoming. To improve the ground truth, a specific label thickness must be set for each individual road object.
Besides, polygonal chain labels do not capture perfectly irregular road borders, which the predictions match more closely. In this specific regard, the models outperform the ground truth in terms of pixel-wise correspondence to the roads, but are yet penalized in the metrics.
Consequently, straightening the predicted roads would make them coincide better with the labels.
In parallel, because of the absence of object awareness in FCNNs, predicted roads are sometimes disconnected at intersections. The next step after road candidate extraction is the construction of a road graph which can be optimized to reconnect loose segments to each other. This is however outside the scope of this letter.

\textbf{Strengths of the method:} The proposed method overcomes the major difficulty of isolating thin objects in a speckled environment and detecting many road patterns despite significant visual differences. The FCNNs were trained using a small dataset, relatively to other datasets used for deep learning. They succeeded nonetheless, not only for the area they were fine-tuned on (Lincoln) but also for other completely unrelated areas (Kalisz and Bonn). FCNNs show an encouraging potential for adaptation given the complexity of the task at hand, and would undoubtedly benefit from further training on additional images.
Moreover, the predictions are smooth, for the most part continuous and almost entirely free of noise, showing that FCNNs successfully leverage the image wide context to improve the consistency of local predictions. The construction of a road graph can therefore be applied without any pre-processing on the road candidates, as they already constitute a solid baseline segmentation.

\section{Conclusion} \label{section:conclusion}
Fully Convolutional Neural Networks (FCNNs) prove to be an effective solution to perform road extraction from SAR images. We establish that off-the-shelf FCNNs can be substantially enhanced specifically for road segmentation by adding a tolerance rule towards spatially small mistakes. Our version of Deeplabv3+ modified with a Mean Squared Error regression loss, rebalanced towards the road class, achieves in average 44\% intersection over union across our test sets. We also show that FCN-8s, no longer the state-of-the-art, reaches scores very close to those of DeepLabv3+, while being far shallower. However FCN-8s' predictions are more noisy and less smooth, making DeepLabv3+ a more robust road candidate extractor. This narrow performance gap points out the need to design new FCNN architectures specialized for road segmentation.
The use of FCNNs as highly adaptable road candidate extractors should provide future works with a reliable means to obtain prior segmentations, on which graph reconstruction can be applied to map entire road networks.

\ifCLASSOPTIONcaptionsoff
  \newpage
\fi

\bibliographystyle{IEEEtran}
\bibliography{bibtex/bib/bibliography.bib}

\end{document}